\documentclass[nohyperref]{article}

\usepackage{microtype}
\usepackage{graphicx}
\usepackage{subfigure}
\usepackage{ bbold }
\usepackage{booktabs} 

\usepackage{hyperref}

\usepackage{icml2022}

\usepackage{amsmath}
\usepackage{amssymb}
\usepackage{mathtools}
\usepackage{amsthm}
\usepackage{float}
\usepackage{booktabs, multirow}
\usepackage[capitalize,noabbrev]{cleveref}

\theoremstyle{plain}

\theoremstyle{definition}

\theoremstyle{remark}

\usepackage[textsize=tiny]{todonotes}

\icmltitlerunning{NCL to PCL with Aggressive Modality Drop}

\begin{document}

\twocolumn[
\icmltitle{Negative Co-learning to Positive Co-learning with Aggressive Modality Drop}



\icmlsetsymbol{equal}{*}


\begin{icmlauthorlist}
\icmlauthor{Nicholas Magal}{equal,cmu}
\icmlauthor{Minh Tran}{equal, cmu}
\icmlauthor{Riku Arakawa}{equal, cmu}
\icmlauthor{Suzanne Nie}{equal, cmu}
\end{icmlauthorlist}

\icmlaffiliation{cmu}{School of Computer Science, Carnegie Mellon University, Pittsburgh, PA, USA}
\icmlcorrespondingauthor{Nicholas Magal}{first1.last1@xxx.edu}
\icmlcorrespondingauthor{Minh Tran}{first2.last2@www.uk}

\icmlkeywords{Machine Learning, ICML}

\vskip 0.3in
]



\printAffiliationsAndNotice{\icmlEqualContribution} 

\begin{abstract}
 This paper aims to document an effective way to improve multimodal co-learning by using aggressive modality dropout. We find that by using aggressive modality dropout we are able to reverse negative co-learning (NCL) to positive co-learning (PCL). Aggressive modality dropout can be used to 'prep' a multimodal model for unimodal deployment, and dramatically increases model performance during negative co-learning, where during some experiments we saw a 20\% gain in accuracy. We also benchmark our modality dropout technique against PCL to show that our modality drop out technique improves co-learning during PCL, although it does not have as much as an substantial effect as it does during NCL.
\end{abstract}

\section{Introduction}
Multimodal Machine Learning can provide more robust and accurate models that are able to obtain higher performance then their unimodal variants \cite{rahate2022multimodal}. However, in many cases multimodal models rely on having high availability to the same modalities during training and test time.  This is not a realistic assumption, as there are many cases where the modalities present during train time are very different from those during test time. For example, it is possible for all modalities to be present only $80\%$ of time during deployment with only $20\%$ of the time having strictly a unimodal modality present \cite{rahate2022multimodal}. In a case where only one model can be deployed, it is therefore essential for the multimodal model to perform acceptably in both the unimodal as well as the multimodal case, ideally performing better then a unimodal model varient. 

Motivated by this problem, we propose a solution for reversing NCL to PCL. We refer to positive co-learning (PCL) as cases where co-learning leads to better performance during test time on uni-modal data compared to unimodally trained models, while negative co-learning (NCL) refers to instances where the multimodally trained model performs worse on test time when compared to the unimodally trained variant. We show that in situations where there is NCL, by applying aggressive modality dropout we are able to reverse NCL to PCL. While there is prior work documenting the effectiveness of modality modality dropout during co-learning and multimodal machine learning, we are the first to show that modality dropout can reverse NCL to PCL. 
 
\section{Related Work}

Modality interactions occur when modalities are integrated together in multi-modal machine learning pipelines \cite{https://doi.org/10.48550/arxiv.2209.03430}.  There are many types of interactions, not all of them positive. For example, given two different modalities one modality can become dominant where one modality can dominate model predictions \cite{https://doi.org/10.48550/arxiv.2201.02184}. Modality dropout is one method to help combat this. Below, we go over several papers that incorporate modality dropout for this use case. We also briefly go over co-learning.

In \cite{https://doi.org/10.48550/arxiv.2201.02184}, the authors create Audio-Visual Hidden Unit BERT (AV-HuBERT). AV-HuBERT is a transformer based representation learning framework for audio-visual representations. The authors note that during training the audio input can dominate model decisions. In order to combat this, modality dropout is employed by assigning probabilities for each modality to be dropped during training. This prevents AV-Hubert from over-relying on one modality.

Another interesting application of modality dropout is used in \cite{https://doi.org/10.48550/arxiv.2005.13616}. In this paper, a talking head is animated using both visual and audio modalities. Typically in generating talking faces from both audio and visual modalities there is such a strong correlation between the video modality and output animation that the model largely ignores the audio input. This results in a degradation of subtle animation components that rely on the audio modality, such as lip-closure animation.  Modality dropout is introduced to force the network to learn from both audio and visual modalities and not over rely on the visual modality. 

Modality Dropout has also been used in audio-visual speech recognition (AVSR). By using both audio and and visual modalities, AVSR systems can perform better than just unimodal systems particularly in situations where the audio signal is very noisy \cite{8682566}. One issue for AVSR systems is that during inference at times the visual modality can be missing. This can severely damages model performance as during training all modalities are usually present. In order to mitigate this issue, the authors of \cite{8682566} use per-frame dropout on the visual input. This preps the model for situations where the visual modality is missing, and increases model robustness.

Additional papers that utilize modality dropout for similar reasons as the above include \cite{https://doi.org/10.48550/arxiv.1911.04890} and \cite{DBLP:journals/corr/NeverovaWTN15}.

\cite{zadeh2020foundations} show the benifits of co-learning by training a Memory Fusion Network (MFN) on multiple modalities that when tested on only one modality outperforms the unimodal version. Specifically, one MFN network is trained on audio, visual, and language and another MFN is only trained on language. During test time, both networks are tested only on language. Surprisingly the multimodel MFN performs better then the unimodal MFN.  This is a case of Weaker Enhancing Stronger (WES), which refers to when weaker modalities are used in training but dropped during test time \cite{rahate2022multimodal}.

\section{Proposed Approach}

\begin{figure} 
    \centering
    \includegraphics[width=0.5\textwidth]{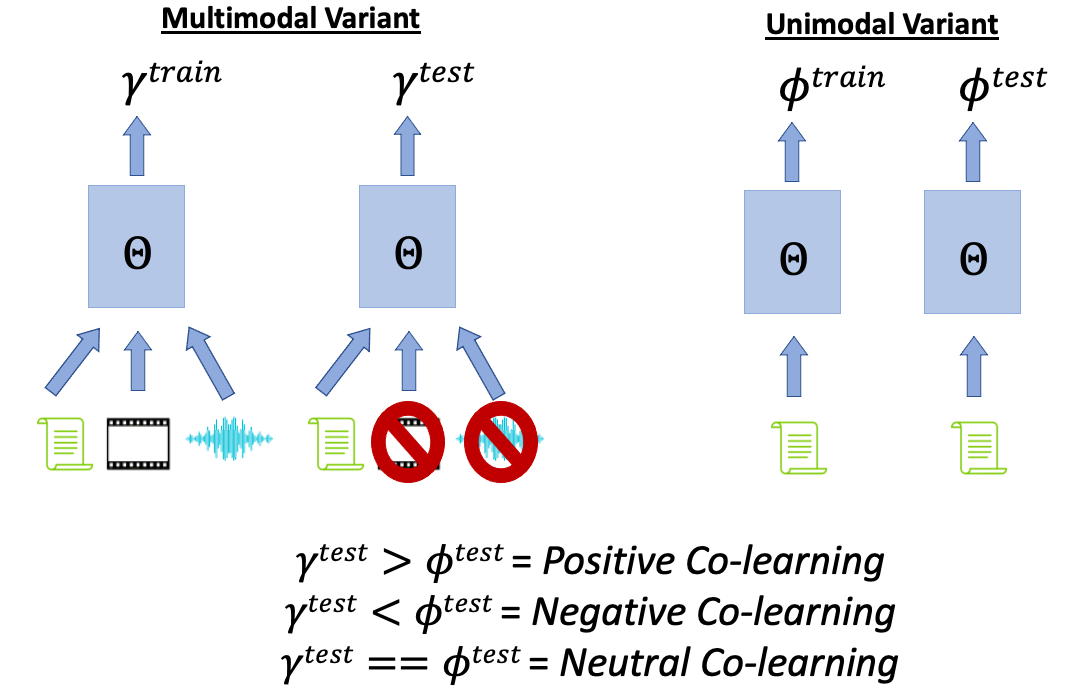}
    \caption{Positive Co-learning (PCL) vs Negative Co-learning (NCL). PCL cases where co-learning leads to better performance during test time on uni-modal data compared to unimodally trained models, while NCL refers to instances where the multimodally trained model performs worse on test time when compared to the unimodally trained variant. Neutral Co-learning (NeCL) refers to cases where both multmodal and unimodal varients perform the same.  }
    \label{fig:majority-label-cm}
\end{figure}

Both a  bidirectional early fusion long short term memory (bi-EFLSTM) as well as a Memory Fusion Network (MFN) are used in our experiments. The bi-EFLSTM model consists of two LSTM cells one moving forward in time and another moving backward in time, followed by two linear layers. Mathematically, a LSTM cell can be represented as:

\begin{small}
\begin{align}
i = \sigma(W_{ii}x + b_{ii} + W_{hi}h + b_{hi})\\
f = \sigma(W_{if}x + b_{if} + W_{hf}h + b_{hf})\\
g = tanh(W_{ig}x + b_{ig} + W_{hg}h + b_{hg})\\
o = \sigma(W_{io}x + b_{io} + W_{ho}h + b_{ho})\\
c' = f *c + i * g
h'=o * tanh(c')
\end{align}
\end{small}

The loss function we use for the bi-EFLSTM is cross entropy which  can be represented as: 
\begin{equation}
H(Y, X) = - \sum_{e \in E} t_e log (p_e)  
\end{equation}

The MFN network consists of three parts: a system of LSTMS for each modality, a Delta-memory Attention Network, and a Multi-view Gated Memory \cite{zadeh2020foundations}. The system of LSTMS are used to model view specific interactions \cite{https://doi.org/10.48550/arxiv.1802.00927}. The Delta-memory Attention Network (DMAN) is used to model cross-view interactions. A coefficient assignment technique is applied on the concatenation of LSTM memories $c^{[t-1,t]}$ to form cross view interactions. These memories are passed to a neural network to obtain the softmax attention scores, which can be represented as:
\begin{equation}
a^{t-1, t} = D_a(c^{[t-1,t]})
\end{equation}
where $D_a$ is a neural network. Following this, we can apply the element wise product to get our final representation:
\begin{equation}
c^{[t-1,t]} = c^{[t-1,t]} \odot a^{t-1, t}
\end{equation}

These results are then stored by using the Multi-view Gated Memory component. The $c^{[t-1,t]}$ we get from our DMAN is passed into our multi view gated memory to signal the important cross-view interactions. A cross view update proposal is then created by inputting $c^{[t-1,t]}$ into a new neural network $D_u$:

\begin{equation}
    u^t = D_u(c^{[t-1,t]})
\end{equation}

Multi-view Gated Memory then uses two gates to decide how much of the proposal to incorporate vs how much of the current memory to remember. These gates are defined as:

\begin{equation}
    \gamma^t_1 = D_{\gamma_1}(c^{[t-1,t]}),\: \gamma^t_2 = D_{\gamma_2}(c^{[t-1,t]}) 
\end{equation}
where $D_{\gamma_1}$ and $D_{\gamma_2}$ are neural networks. Finally, our $u^t$ is updated by:
\begin{equation}
u^t =\gamma^t_1 \odot u^{t-1} + \gamma^t_2 \odot tanh(u^t)
\end{equation}

The loss function we use for our MFN network is the L1 Loss, which can be formulated as:

\begin{equation}
l = 1/n\sum_{n} |x_n-y_n|
\end{equation}

\section{Experimental Setup}

We used both IEMOCAP and MOSI for our experiments. MOSI was forced aligned using P2FA \cite{MFN}. For MOSI, the language was embedded into Glove word embeddings \cite{glove}, audio features were extracted using COVAREP, and the visual features were extracted using Facet. We used the pre processed data-set that was available from \cite{MFN}

We manually pre processed the IEMOCAP dataset. We dropped all data points belonging to 'xxx' and 'other'.  All modalities were forced aligned to the word level using Sphinx III \cite{iemocap}. We also embedded our language using Glove \cite{glove}. We evaluated the distribution of the sentence lengths and found the average length was 13 with a standard deviation of 11, with a max sentence length of 112. In order to account for this imbalance we only included sentences that were within 2 standard deviations of the mean which was anything less then 30 tokens. For sentences that were less then 30 tokens we applied zero padding onto the beginning of the sentence to make each sentence of length 30. The final dimension size for our language dataset was  (6877 x 30 x 300), where 6877 corresponded to our total data points, 30 to the length of each sentence, and 300 to the GLOVE embedding size.

\begin{figure}[!htbp]
  \centering
  \includegraphics[width=0.15\textwidth]{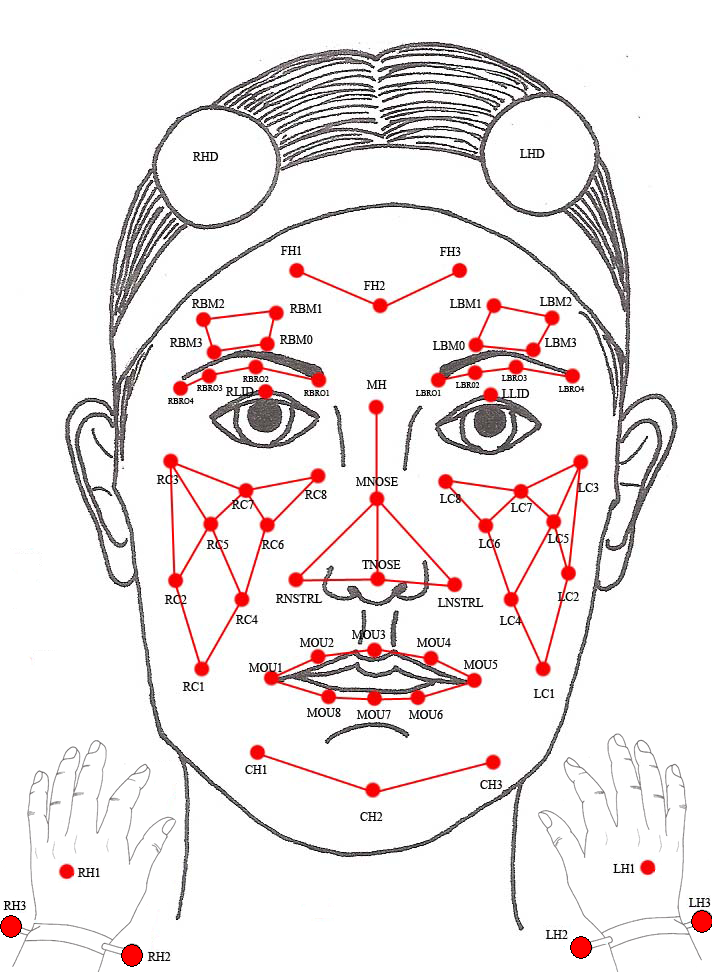}  
  \caption{Extracted Features of Hands and Head in IEMOCAP}
  \label{fig:eimocap}
\end{figure}

For audio, we extracted Mel-frequency cepstral coefficients signals and Melspectrogram arrays. This resulted in a feature embedding of dimension (6877 x 30 x 80).

For the visual component, since context of the background scene remains unchanged through each utterance we cropped the outer frame out and left the middle of the image with the two main subjects. We used sparse sampling \cite{sparse} which preserves information while reduces computational cost significantly. For each processed frame, we extracted features from face detection, facial unit action, and emotion detection and concatenated all the features together into a visual feature vector. The visual feature dimension was (6877 x 30 x 310).

\begin{figure}[!htbp]
    \includegraphics[width=.22\textwidth]{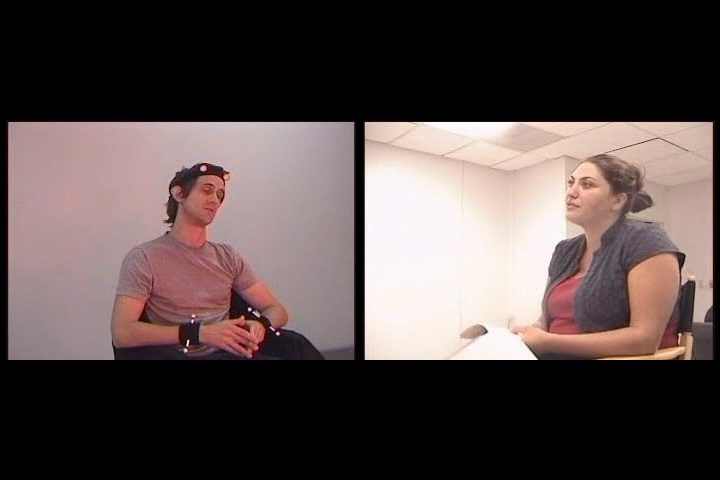}\hfill
    \includegraphics[width=.22\textwidth]{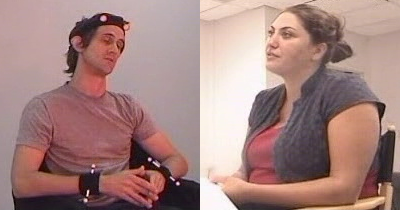}\hfill
    \includegraphics[width=.22\textwidth]{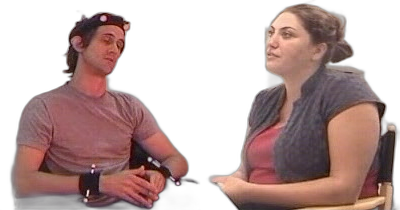}\hfill
    \includegraphics[width=.22\textwidth]{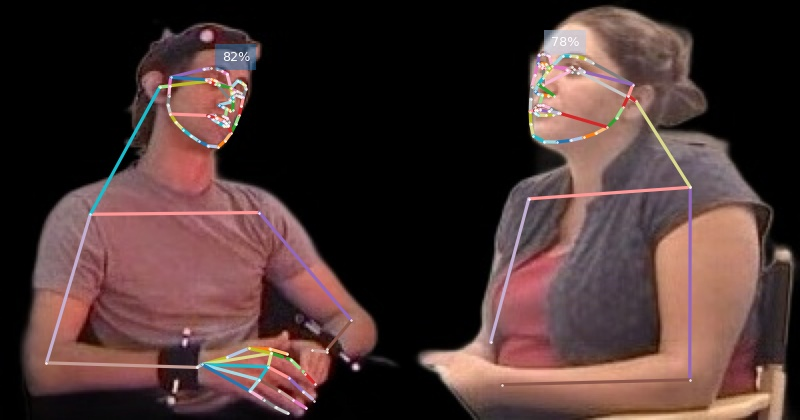}
    \caption{Image Processing}\label{fig:foobar}
\end{figure}

The final multimodal edition of the data concatenates the language, video, and audio modalities together and results in a (6877 x 30 x 690) data set.

We created two different scenarios to measure the impact of modality drop on PCL and NCL. PCL was accomplished by following \cite{zadeh2020foundations} to train a MFN on MOSI that performed better then its unimodal variant. NCL was accomplished by training a bi-EFLSTM that was trained on IEMOCAP. IEMOCAP was chosen due to the increased complexity of the data set that stems from its dyadic conversation format.  We found that this increased complexity alongside the small network complexity of the bi-EFLSTM made it harder for the model to have PCL. 

For both settings, a multimodal network was trained on audio, visual, and language and a unimodal version of the same network was only trained on language. During test time, both networks
were tested only on language. As a baseline, we trained these models with no modality dropout. Afterwords we ran the same experiments except with modality dropout in variable levels to be able to observe the impact of modality dropout.  

Modality dropout can be formulated as: 

\begin{small} 
\begin{equation}
  f^a_t =
    \begin{cases}
      f^a_t & 1-p^a\\
      0 & p^a
    \end{cases}       
\end{equation}

\begin{equation}
  f^l_t =
    \begin{cases}
      f^l_t & 1-p^l\\
      0 & p^l
    \end{cases}       
\end{equation}

\begin{equation}
  f^v_t =
    \begin{cases}
      f^v_t & 1-p^v\\
      0 & p^v
    \end{cases}       
\end{equation}
\end{small}

where $f^a_t,\: f^l_t,\: f^v_t $ are the corresponding audio, language, and video feature vectors at time $t$, and $p^a, \: p^l, \: p_v $ are the probabilities for the audio, language, and video features to be masked out to 0. This process occurs for each batch in the training loop.

We trained the bi-EFLSTM with a learning rate of .0001, batch size of 15, and for 40 epochs. Early stopping was used on all runs according to validation loss on a unimodal validation dataset. Adam was used as our optimizer. We also used a ReduceLRonPlateau for our scheduler. We used 128 as the size of our hidden dimensions. 

We follow the hyper-parameters that observed the highest performance for the MFN network from \cite{https://doi.org/10.48550/arxiv.1802.00927}.

For each experiment, we run 5 different random seeds and report the average results. 
\section{Results and Discussion}

\begin{figure}[h]%
    \centering
{{\includegraphics[width=7cm]{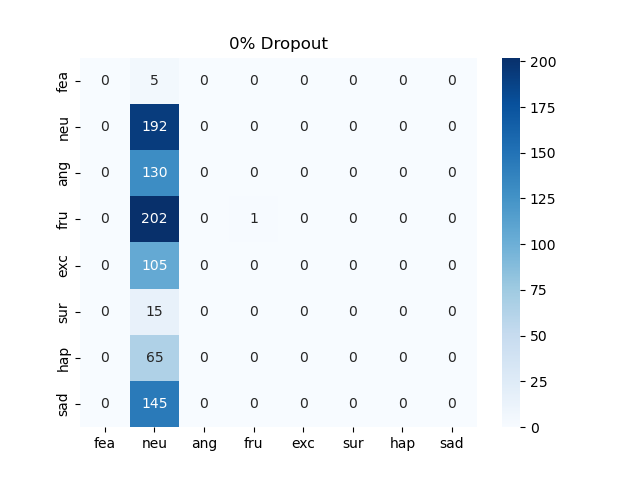} }}%
    \qquad
{{\includegraphics[width=7cm]{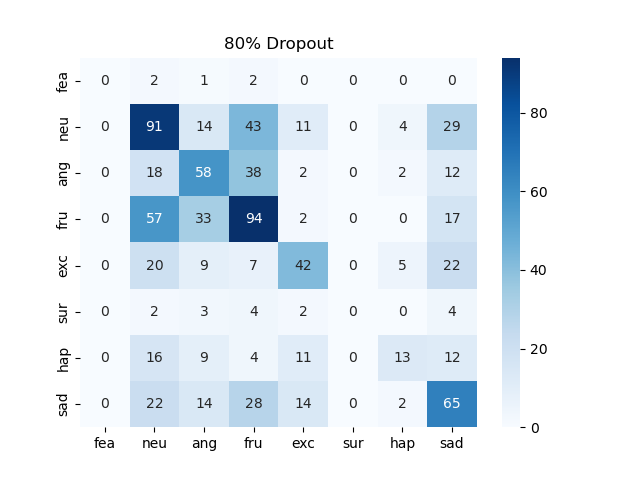} }}%
    \caption{Confusion Matrix of the bi-EFLSTM at 0\% modality dropout vs 80\% dropout on both the audio and visual modalities. As shown, without modality dropout the model struggles to learn anything meaningful and only outputs the neutral class. With modality dropout the model is able to correctly classify other classes and obtains much better performance.}%
    \label{fig:IEMO confusion matrix}%
\end{figure}
From our experiments on the bi-EFLSTM using the IEMOCAP dataset we found that aggressive modality dropout can reverse NCL to PCL. Without modality dropout, our multimodal bi-EFLSTM modal scored only 27\% during test time on the unimodal language data, while the unimodal version scored 45\%, a dramatically higher score. By employing modality drop out with a dropout level of 80\% on the audio and video features on the multimodal bi-EFLSTM during training we are able to score 47\% on the unimodal test dataset, beating our unimodal variant, effectivly reversing NCL to PCL across all metrics.  Not  unexpectedly, performance degrades when increasing modality drop out above 80\%. We speculate that this is due to instability that arises from the model not being given enough time to learn from the other modalities.

\begin{figure*}[h]%
    \centering
{{\includegraphics[width=7cm]{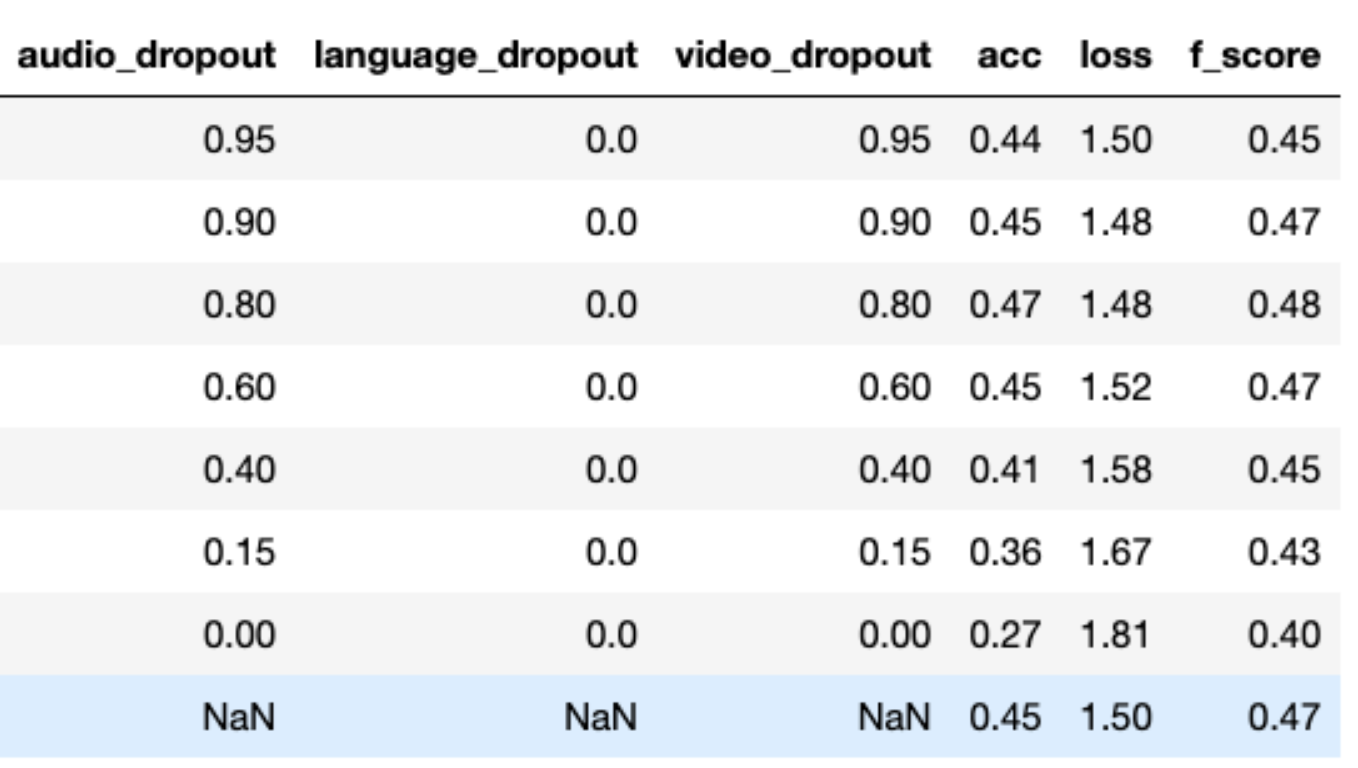} }}%
{{\includegraphics[width=7cm]{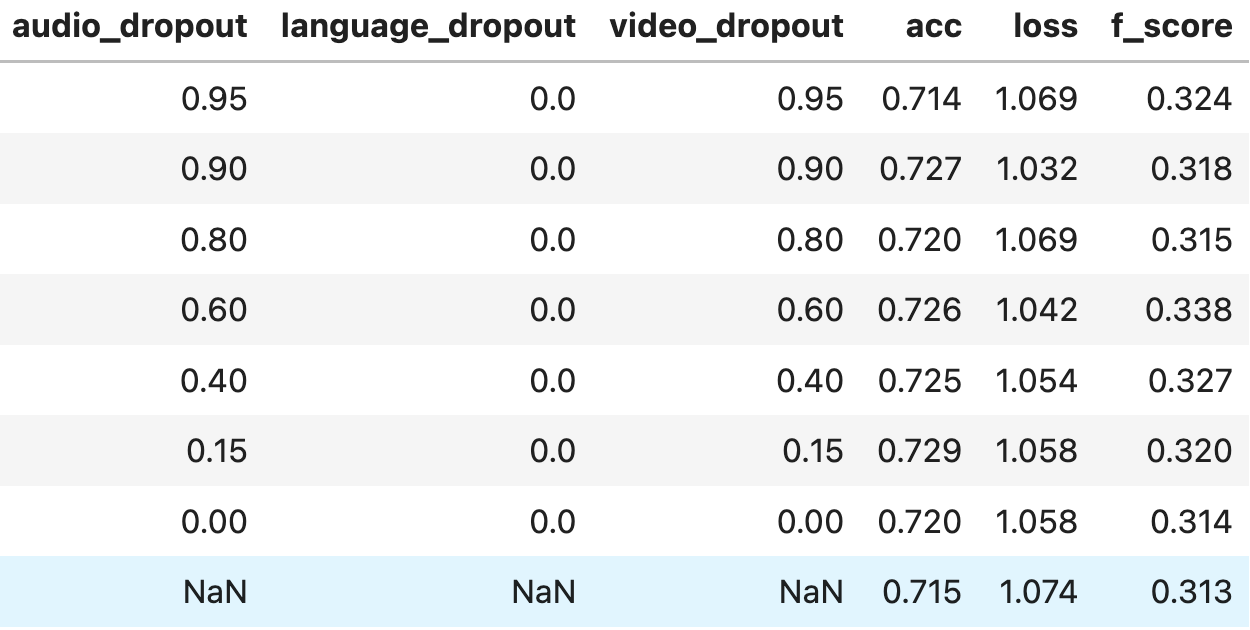} }}%
{{\includegraphics[width=7cm]{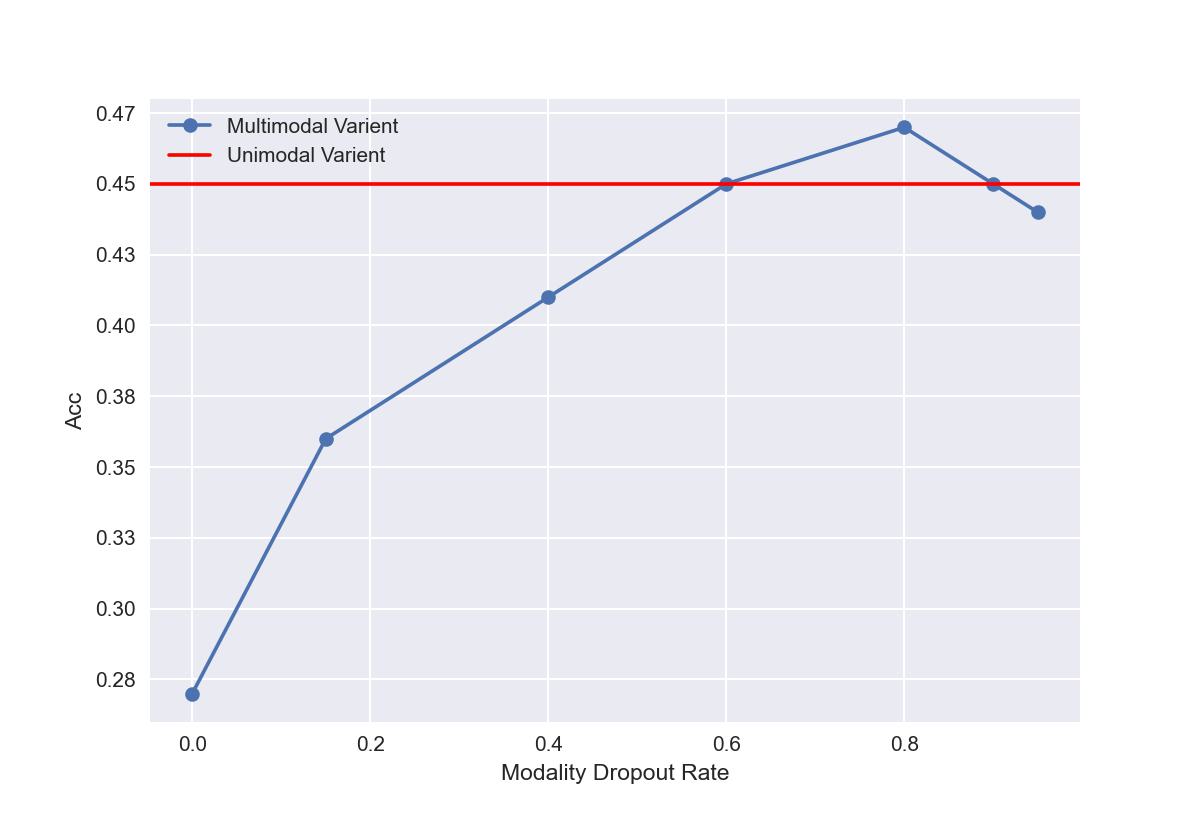} }}%
{{\includegraphics[width=7cm]{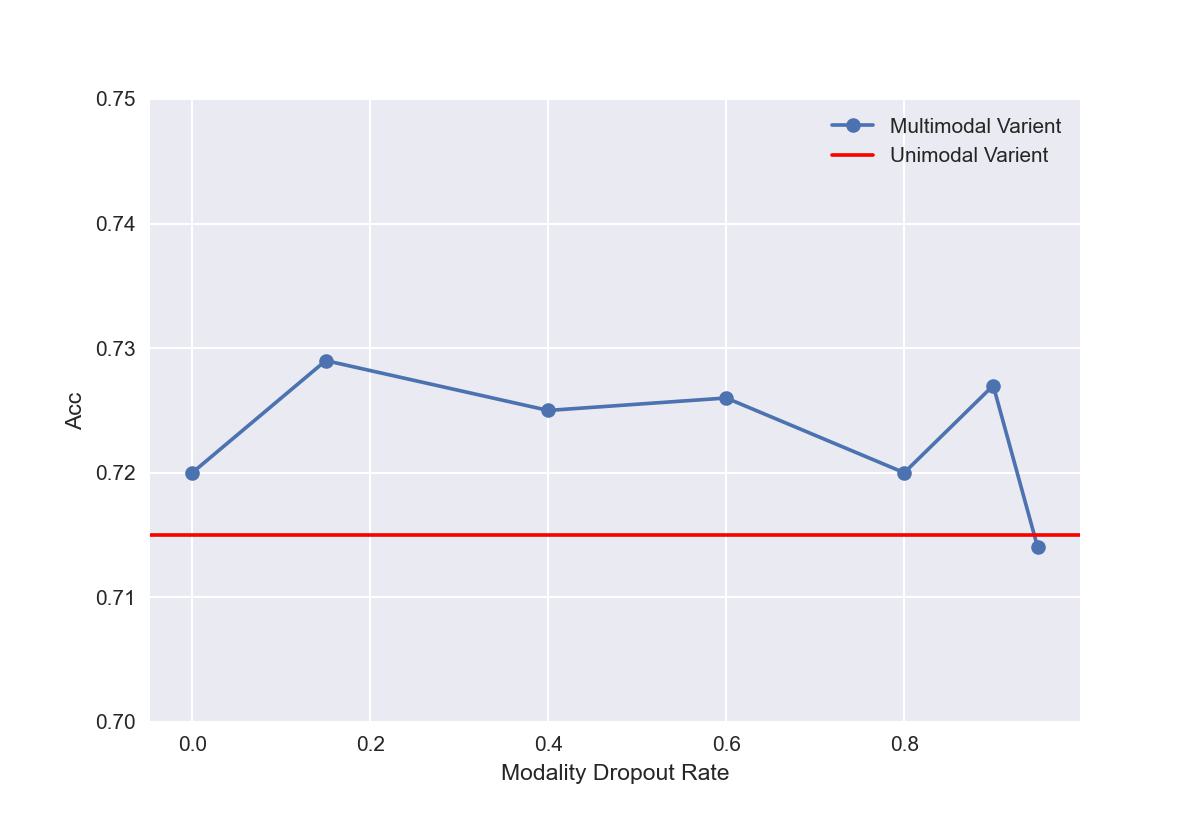} }}%
    \caption{Performance using the bi-EFLSTM on Iemocap (left) using modality drop out and using the MFN on MOSI (right). The light blue row on the table figures corresponds to the unimodal variant, with the rest of the rows being the multimodal variant. As shown, modality dropout using the bi-EFLSTM substantially improves the co-learning process across all metrics, and is able to reverse PCL to NCL.  Modality drop out during PCL using the MFN
does not have as big as an effect as it does during NCL. However, modality dropout still improves metrics compared to
non dropout settings.
}%
    \label{bi-EFLSTM performance}%
\end{figure*}

Although the effects were not as significant as the results we found on our bi-EFLSTM experiments, we also found that co-learning with modality dropout on our MFN network that already had achieved PCL had positive benefits from training with modality dropout. Without modality dropout, our multimodal MFN achieved 1.058 MAE, 72\% accuracy, and .314 f-score on the unimodal test data while the unimodal variant scored 1.074 MAE, 71.5\% accuracy, and 0.313 f-score. When we introduced modality dropout, our PCL was improved slightly, with all metrics increasing slightly across different levels of modality dropout. 

In cases NCL occurs, aggressive modality drop is shown to significantly improve the co-learning process. We hypothesis that in cases where NCL is present, the model overemphasizes importance of the supporting modalities and therefore collapses when they are not present. Figure 6 confirms this hypothesis, and shows where no modality dropout is present the model only chooses one output class for its predictions.  To force the the model to not overemphasize these modalities, modality dropout can be applied to force the model to be able to perform well in cases where the supporting modalities are no longer present. In cases where PCL occurs, our empirical results show that modality drop out is not as important. This is probably due to the model implicitly not overemphasizing supporting modalities. However, as shown, there are still some benefits to using modality dropout during PCL, suggesting that even in these cases there is some overemphasis.

And finally, training with modality dropout during co-learning intuitively makes sense. If during test time we are going to be testing on unimodal data, it makes sense to "prep" the model during training by presenting a situation similar to training where only one modality is present. By using modality drop out, we capture a hybrid of both unimodal and multimodal training, where the sum result can be a more powerful and robust model.

\section{Conclusion and Future Directions}

We present a novel application of modality drop and show its effects on NCL and PCL. Future work can be extended to show how this effect changes based off of a partial modality drop, instead of the full modality drop we presented in this paper. Also, it is worth investigating how modality drop during co-learning is impacted by choosing different primary modalities for the co-learning process. And as a final future direction, we aim to investigate finding a modality dropout level that works for both multimodal inference as well as unimodal inference. 


\clearpage
\bibliography{11777}

\begin{thebibliography}{13}
\providecommand{\natexlab}[1]{#1}
\providecommand{\url}[1]{\texttt{#1}}
\expandafter\ifx\csname urlstyle\endcsname\relax
  \providecommand{\doi}[1]{doi: #1}\else
  \providecommand{\doi}{doi: \begingroup \urlstyle{rm}\Url}\fi

\bibitem[Abdelaziz et~al.(2020)Abdelaziz, Theobald, Dixon, Knothe, Apostoloff,
  and Kajareker]{https://doi.org/10.48550/arxiv.2005.13616}
Abdelaziz, A.~H., Theobald, B.-J., Dixon, P., Knothe, R., Apostoloff, N., and
  Kajareker, S.
\newblock Modality dropout for improved performance-driven talking faces, 2020.
\newblock URL \url{https://arxiv.org/abs/2005.13616}.

\bibitem[Busso et~al.(2008)Busso, Bulut, Lee, Kazemzadeh, Mower, Kim, Chang,
  Lee, and Narayanan]{iemocap}
Busso, C., Bulut, M., Lee, C.-C., Kazemzadeh, A., Mower, E., Kim, S., Chang,
  J.~N., Lee, S., and Narayanan, S.~S.
\newblock Iemocap: interactive emotional dyadic motion capture database.
\newblock \emph{Language Resources and Evaluation}, 42\penalty0 (4):\penalty0
  335--359, 2008.
\newblock \doi{10.1007/s10579-008-9076-6}.
\newblock URL \url{https://doi.org/10.1007/s10579-008-9076-6}.

\bibitem[Lei et~al.(2021)Lei, Li, Zhou, Gan, Berg, Bansal, and Liu]{sparse}
Lei, J., Li, L., Zhou, L., Gan, Z., Berg, T.~L., Bansal, M., and Liu, J.
\newblock Less is more: Clipbert for video-and-language learning via sparse
  sampling, 2021.
\newblock URL \url{https://arxiv.org/abs/2102.06183}.

\bibitem[Liang et~al.(2022)Liang, Zadeh, and
  Morency]{https://doi.org/10.48550/arxiv.2209.03430}
Liang, P.~P., Zadeh, A., and Morency, L.-P.
\newblock Foundations and recent trends in multimodal machine learning:
  Principles, challenges, and open questions, 2022.
\newblock URL \url{https://arxiv.org/abs/2209.03430}.

\bibitem[Makino et~al.(2019)Makino, Liao, Assael, Shillingford, Garcia, Braga,
  and Siohan]{https://doi.org/10.48550/arxiv.1911.04890}
Makino, T., Liao, H., Assael, Y., Shillingford, B., Garcia, B., Braga, O., and
  Siohan, O.
\newblock Recurrent neural network transducer for audio-visual speech
  recognition, 2019.
\newblock URL \url{https://arxiv.org/abs/1911.04890}.

\bibitem[Neverova et~al.(2015)Neverova, Wolf, Taylor, and
  Nebout]{DBLP:journals/corr/NeverovaWTN15}
Neverova, N., Wolf, C., Taylor, G.~W., and Nebout, F.
\newblock Moddrop: adaptive multi-modal gesture recognition.
\newblock \emph{CoRR}, abs/1501.00102, 2015.
\newblock URL \url{http://arxiv.org/abs/1501.00102}.

\bibitem[Pennington et~al.(2014)Pennington, Socher, and Manning]{glove}
Pennington, J., Socher, R., and Manning, C.~D.
\newblock Glove: Global vectors for word representation.
\newblock In \emph{Empirical Methods in Natural Language Processing (EMNLP)},
  pp.\  1532--1543, 2014.
\newblock URL \url{http://www.aclweb.org/anthology/D14-1162}.

\bibitem[Rahate et~al.(2022)Rahate, Walambe, Ramanna, and
  Kotecha]{rahate2022multimodal}
Rahate, A., Walambe, R., Ramanna, S., and Kotecha, K.
\newblock Multimodal co-learning: challenges, applications with datasets,
  recent advances and future directions.
\newblock \emph{Information Fusion}, 81:\penalty0 203--239, 2022.

\bibitem[Shi et~al.(2022)Shi, Hsu, Lakhotia, and
  Mohamed]{https://doi.org/10.48550/arxiv.2201.02184}
Shi, B., Hsu, W.-N., Lakhotia, K., and Mohamed, A.
\newblock Learning audio-visual speech representation by masked multimodal
  cluster prediction, 2022.
\newblock URL \url{https://arxiv.org/abs/2201.02184}.

\bibitem[Zadeh et~al.(2018{\natexlab{a}})Zadeh, Liang, Mazumder, Poria,
  Cambria, and Morency]{MFN}
Zadeh, A., Liang, P.~P., Mazumder, N., Poria, S., Cambria, E., and Morency,
  L.-P.
\newblock Memory fusion network for multi-view sequential learning,
  2018{\natexlab{a}}.
\newblock URL \url{https://arxiv.org/abs/1802.00927}.

\bibitem[Zadeh et~al.(2018{\natexlab{b}})Zadeh, Liang, Mazumder, Poria,
  Cambria, and Morency]{https://doi.org/10.48550/arxiv.1802.00927}
Zadeh, A., Liang, P.~P., Mazumder, N., Poria, S., Cambria, E., and Morency,
  L.-P.
\newblock Memory fusion network for multi-view sequential learning,
  2018{\natexlab{b}}.
\newblock URL \url{https://arxiv.org/abs/1802.00927}.

\bibitem[Zadeh et~al.(2020)Zadeh, Liang, and Morency]{zadeh2020foundations}
Zadeh, A., Liang, P.~P., and Morency, L.-P.
\newblock Foundations of multimodal co-learning.
\newblock \emph{Information Fusion}, 64:\penalty0 188--193, 2020.

\bibitem[Zhang et~al.(2019)Zhang, Lei, Ma, and Xie]{8682566}
Zhang, S., Lei, M., Ma, B., and Xie, L.
\newblock Robust audio-visual speech recognition using bimodal dfsmn with
  multi-condition training and dropout regularization.
\newblock In \emph{ICASSP 2019 - 2019 IEEE International Conference on
  Acoustics, Speech and Signal Processing (ICASSP)}, pp.\  6570--6574, 2019.
\newblock \doi{10.1109/ICASSP.2019.8682566}.

\end{thebibliography}
\bibliographystyle{icml2022}

\newpage
\appendix


\end{document}